\documentclass{article} 
\usepackage{nips15submit_e,times}
\usepackage{hyperref}
\usepackage{url}

\usepackage{amsmath}
\usepackage{amssymb}
\usepackage{caption}
\usepackage{amsthm}

\usepackage{xcolor}
\usepackage{graphicx}
\hypersetup{
    colorlinks,
    linkcolor={red!50!black},
    citecolor={blue!50!black},
    urlcolor={blue!80!black}
}

\title{Macau: Scalable Bayesian Multi-relational Factorization with Side Information using MCMC}

\author{
Jaak Simm\thanks{Adam Arany and Jaak Simm contributed both equally as first authors.} \\
ESAT-STADIUS, KU Leuven\\
B-3001 Heverlee, Belgium \\
\texttt{jsimm@esat.kuleuven.be} \\
\And
Adam Arany\footnotemark[1] \\
ESAT-STADIUS, KU Leuven\\
B-3001 Heverlee, Belgium \\
\texttt{aarany@esat.kuleuven.be} \\
\AND
Pooya Zakeri \\
ESAT-STADIUS, KU Leuven\\
B-3001 Heverlee, Belgium \\
\texttt{pzakeri@esat.kuleuven.be} \\
\And
Tom Haber \\
Hasselt University \\
3500 Hasselt, Belgium \\
\texttt{tom.haber@uhasselt.be} \\
\And
J\"org K. Wegner \\
Janssen Pharmaceutica \\
B-2340 Beerse, Belgium \\
\texttt{jwegner@its.jnj.com} \\
\And
Vladimir Chupakhin \\
Janssen Pharmaceutica \\
B-2340 Beerse, Belgium \\
Janssen Pharmaceuticals \\
45007 Toledo, Spain\\
\texttt{vchupakh@its.jnj.com} \\
\And
Hugo Ceulemans \\
Janssen Pharmaceutica \\
B-2340 Beerse, Belgium \\
\texttt{hceulema@its.jnj.com} \\
\And
Yves Moreau \\
ESAT-STADIUS, KU Leuven \\
B-3001 Heverlee, Belgium \\
\texttt{moreau@esat.kuleuven.be} \\
}

\newcommand{\Ub}{\mathbf{U}}
\newcommand{\Xb}{\mathbf{X}}
\newcommand{\Eb}{\mathbf{E}}
\newcommand{\ub}{\mathbf{u}}
\newcommand{\vb}{\mathbf{v}}

\newcommand{\jb}{\mathbf{j}}
\newcommand{\xb}{\mathbf{x}}
\newcommand{\alphab}{\boldsymbol{\alpha}}
\newcommand{\nlatent}{D}     
\newcommand{\IR}{I_R}        
\newcommand{\lmean}{\boldsymbol{\mu}}     
\newcommand{\lprec}{\Lambda} 

\newcommand{\fweightall}{\beta}
\newcommand{\fweightregall}{\lambda}
\newcommand{\fweight}[1]{\beta_{#1}}
\newcommand{\fweighthat}[1]{\hat{\beta}_{#1}}
\newcommand{\fweightreg}[1]{\lambda_{\beta_{#1}}}

\newcommand{\feat}[2][u]{\mathbf{x}^{(#1)}_{#2}}
\newcommand{\latent}[2]{\ub^{(#1)}_{#2}}

\newcommand{\eset}{\mathcal{E}}
\newcommand{\rset}{\mathcal{R}}

\newcommand{\Kb}{\mathbf{K}}
\newcommand{\Lambdab}{\boldsymbol{\Lambda}}

\newtheorem{theorem}{Theorem}[section]
\newtheorem{lemma}[theorem]{Lemma}

\DeclareMathOperator{\vecop}{vec}
\DeclareMathOperator{\tr}{tr}

\nipsfinalcopy

\begin{document}

\maketitle

\begin{abstract}
We propose Macau, a powerful and flexible Bayesian factorization method for heterogeneous data.
Our model can factorize any set of entities and relations that can be represented by a relational model, 
including tensors and also multiple relations for each entity.
Macau can also incorporate side information, specifically entity and relation features, which are crucial for predicting sparsely observed relations.
Macau scales to millions of entity instances, hundred millions of observations, and sparse entity features with millions of dimensions.
To achieve the scale up, we specially designed sampling procedure for entity and relation features that relies primarily on noise injection in linear regressions.
We show performance and advanced features of Macau in a set of experiments, including challenging drug-protein activity prediction task.

\end{abstract}

\section{Introduction}
\emph{Matrix factorization} (MF) has a long history and a wide range of applications in data sciences, engineering and many fields of scientific research.
While classical approaches, such as SVD, factorize fully observed matrices, a previous work proposed matrix factorization for partially observed matrices \cite{srebro2003weighted}.
This enabled the direct use of MF in predictive machine learning problems (\emph{e.g.}, in collaborative filtering).
However, the original formulation \cite{srebro2003weighted} can easily overfit the data and it was improved in Probabilistic Matrix Factorization (PMF) \cite{mnih2007probabilistic}.
The main idea in these two papers and in subsequent research has been to represent each row and each column by a latent vector of size $\nlatent$ and find the best match to the observed elements of the matrix:
\begin{equation}
  \label{eq:mf-basic}
  \min_{\ub,\vb} \sum_{(i,j) \in \IR} (R_{ij} - \ub_i ^ \top \vb_j)^2
               + \lambda_{u} \| \ub \|_F^2
               + \lambda_{v} \| \vb \|_F^2,
\end{equation}
where $\ub_i, \vb_j \in \mathbb{R}^\nlatent$ are the latent vectors for $i$th row and $j$th column, $\IR$ is the set of matrix cells whose value has been observed, $R_{ij} \in \mathbb{R}$ are the observed values, $\|\cdot\|_F$ is the Frobenius norm, and $\lambda_{u}, \lambda_{v} > 0$ are regularization parameters.
The last two terms in optimization problem \eqref{eq:mf-basic}, introduced in PMF \cite{mnih2007probabilistic}, are derived from zero-mean Gaussian priors on $\ub_i$ and $\vb_j$ and a Gaussian noise model on the observed values $R_{ij}$.

\subsection{Bayesian PMF}
Bayesian PMF (BPMF) \cite{salakhutdinov2008bayesian} extends PMF to full Bayesian inference by introducing common multivariate Gaussian priors for the latent variables, one for the rows and one for the columns.
To infer these two priors from the data, BPMF places fixed uninformative Normal-Wishart hyperpriors on them.
Let $\lmean_u$ and $\lprec_u$ ($\lmean_v$ and $\lprec_v$) be the mean and precision matrix of the Gaussian prior for rows (columns) then the model used by BPMF is
\begin{align}
  \label{eq:bpmf-latent-u}
  p(\ub,\lmean_u,\lprec_u | \Theta_0) &= \prod_{i=1} \mathcal{N}(\ub_i | \lmean_u, \lprec_u^{-1}) \mathcal{NW}(\lmean_u, \lprec_u | \Theta_0) \\
  \label{eq:bpmf-latent-v}
  p(\vb,\lmean_v,\lprec_v | \Theta_0) &= \prod_{j=1} \mathcal{N}(\vb_j | \lmean_v, \lprec_v^{-1}) \mathcal{NW}(\lmean_v, \lprec_v | \Theta_0),
\end{align}
where $\mathcal{N}$ and $\mathcal{NW}$ are Normal and Normal-Wishart distributions and $\Theta_0$ are the fixed hyperparameters of the Normal-Wishart hyperprior.
Similarly to PMF the noise model of BPMF is Gaussian:
\begin{equation}
  \label{eq:bpmf-noise}
  p(R | \ub, \vb, \alpha_R) = \prod_{(i,j) \in I_R} \mathcal{N}(R_{ij} | \ub_i ^ \top \vb_j, \alpha_R^{-1}),
\end{equation}
where $\alpha_R > 0$ is the precision parameter and is assumed to be known.
From \eqref{eq:bpmf-latent-u}--\eqref{eq:bpmf-noise}, it is straight-forward to derive block Gibbs sampler for each latent vector $\ub_i$ and $\vb_j$, and for the parameters of the Gaussian priors $\lmean_u$, $\lprec_u$, $\lmean_v$, $\lprec_v$.

It is generally observed that BPMF shows improvement in predictive performance compared to PMF (\emph{e.g.}, in the collaborative filtering task of Netflix \cite{salakhutdinov2008bayesian}).
Another advantage of BPMF is that it provides credibility intervals for the estimates.
It should be also noted that BPMF can be easily parallelized and can handle large scale data sets, such as the Netflix challenge data, which contains 200 million observations.

\subsection{Proposed Method}
In this paper we propose Macau, a powerful and flexible method for factorization of heterogeneous data. Its essential features are
\begin{itemize}
 \item The ability to factorize wide range of data models, which we represent by a hypergraph where entities are nodes and relations are hyperedges. Supported models include the cases of ordinary graphs and tensor relations, see Appendix~\ref{appendix:data-models}.
 \item The incorporation of features (side information) for any entity and for any relation.
 \item Scalability up to millions of entity instances, hundred millions of observations, and sparse entity features with millions of dimensions.
\end{itemize}

We follow the approach of BPMF by proposing a Gibbs sampler scheme that includes specially designed 
\emph{noise injection} step for entity and relation features.
This enables the scaling of the method to millions of sparse or to tens of thousands of dense features.

The \emph{novelty} of the proposed method is the combination of all above mentioned functionality into a unified Bayesian framework (see Section~\ref{sec:related-research} for detail overview).
Also in the context of matrix factorization with entity features Macau carries out \emph{MCMC inference} rather than variational approximate approaches, such as Variational Bayes as proposed in previous research \cite{park2013hierarchical}.

We apply our method to a standard matrix factorization benchmark of MovieLens, outperforming the state-of-the-art MF approaches.
Additionally, we explore the performance of Macau in a challenging biochemistry task of drug-protein activity prediction where we demonstrate the effectiveness of the aforementioned characteristics of the method.
This task is based on publicly available data from ChEMBL \cite{bento2014chembl}.
Finally, we report runtime information for private industrial data set from 						 Pharmaceutica containing millions of drug candidates with millions of sparse features and tens of millions of observed activity values.

Our contribution includes an open source package\footnote{URL to the package: \url{https://github.com/jaak-s/BayesianDataFusion.jl}} implementing all of the above mentioned features together with multi-core and multi-node parallelization in the Julia language.

\section{Macau}
\label{sec:macau}
In this section we outline the probabilistic model for Macau.
Then we give an overview of related research (Section~\ref{sec:related-research}).
Finally, we outline the details for the Gibbs sampler (Section~\ref{sec:gibbs}), including the crucial noise injection based scheme for sampling the weight variables linking the entity and relation features (Section~\ref{sec:macau-noise-injection}).

\subsection{Multiple relations and tensor relations}
In practice, data sets can often contain multiple relations between entities (\emph{e.g.}, drugs and proteins, see Fig.~\ref{fig:tensor-model}). 
To handle it in Macau, we consider a relational model with a set of entities $\eset$ and a set of relations $\rset$ such that each relation $R \in \rset$ can link together two or more entities, i.e., $R$ is a tensor.
Each relation $R$ maps the instances of its entities to a real number, denoted by $R_\jb$ where $\jb = (j_1, \ldots, j_k)$ is the index vector and $k$ is the degree of the relation (\emph{i.e.}, the number of entities connected by $R$).
Formally, $R$ is a map $\mathbb{N}^k \rightarrow \mathbb{R}$.
As in the case of partially observed matrix the values $R_{j_1, \ldots, j_k}$ are partially observed.
We denote the latent vector of instance $i \in \mathbb{N}$ of entity $e \in \eset$ by $\latent{e}{i} \in \mathbb{R}^\nlatent$.

Each relation $R$ has a Gaussian noise model with precision $\alpha_R > 0$
\begin{equation}
  \label{eq:macau-noise}
  p(R | \ub, \alpha_R) = \prod_{\jb \in I_R} \mathcal{N}(R_\jb | \mathbf{1} ^ \top \ub_\jb ^ {\eset_R}, \alpha_R^{-1}),
\end{equation}
where $\IR \subset \mathbb{N}^k$ is the set of index vectors for which $R$ is observed, $\eset_R$ is the ordered list of entities connected by relation $R$ where the order is the same as in the index vectors $\jb \in \IR$, $\mathbf{1}$ is the vector of ones and $\ub_\jb ^ {\mathbf{e}} = \ub^{(e_1)}_{j_1} \circ \ldots \circ \ub^{(e_k)}_{j_k}$ is the element-wise product of the latent vectors.
The conditional probability of the observations of all relations is then
\begin{equation}
  \label{eq:macau-noise-full}
  p(\rset | \ub, \alphab) = \prod_{R \in \rset} \prod_{\jb \in I_R} \mathcal{N}(R_\jb | \mathbf{1} ^ \top \ub_\jb ^ {\eset_R}, \alpha_R^{-1}).
\end{equation}
Equation~\eqref{eq:macau-noise-full} allows Macau to simultaneously factorize more than two entities and multiple relations with possibly different degrees.

\subsection{Entity Features}
Entity features are extra information available about instances of entities, often referred to as side-information.
For example, in the case of movie ratings, it could be the genre and the release year for movies, or the age and the gender for users.
In the example of drug-protein activity modeling, it is possible to use substructure information of the drug candidate, represented by a large sparse binary vector.
The idea exploited in Macau is that we can use this extra information to predict the latent vector of the instance and, thus, get more accurate factorization, especially for entities that have few or no observations.

First, let us write the standard Gaussian prior \eqref{eq:bpmf-latent-u}, used in BPMF, for the latent variable $\latent{e}{i}$ of an instance $i$ of entity $e$:
\begin{equation}
  \label{eq:macau-latent-no-features}
  p(\latent{e}{i} | \lmean_e, \lprec_e) = \mathcal{N}(\latent{e}{i} | \lmean_e, \lprec_e^{-1}),
\end{equation}
where $\lmean_e$ and $\lprec_e$ are the common prior mean and precision matrix for entity $e$, respectively.
To incorporate the instance's feature $\feat[e]{i} \in \mathbb{R}^{F_e}$ we add a term $\fweight{e}^\top \feat[e]{i}$ into the Gaussian mean:
\begin{equation}
  \label{eq:macau-latent-with-features}
  p(\latent{e}{i} | \feat[e]{i}, \lmean_e, \lprec_e) =
    \mathcal{N}(\latent{e}{i} | \lmean_e + \fweight{e}^\top \feat[e]{i}, \lprec_e^{-1}),
\end{equation}
where $\fweight{e} \in \mathbb{R}^{F_e \times \nlatent}$ is the weight matrix for the entity features and $F_e$ is the dimensionality of the features.
Equation \eqref{eq:macau-latent-with-features} can be interpreted as a linear model for the latent vectors.
If an instance does not have any observations then the distribution of its latent variable is fully determined by \eqref{eq:macau-latent-with-features}, because there are no terms involving its latent variable in \eqref{eq:macau-noise-full}.
On the other hand, if the instance has many observations its features will have only a minor impact.

To have a full Bayesian treatment for $\fweight{e}$, we introduce a zero mean multivariate normal as its prior:
\begin{align}
  \label{eq:macau-betareg}
  p(\fweight{e} | \lprec_e, \fweightreg{e})
    & = \mathcal{N}(\vecop(\beta_e) | \mathbf{0}, \lprec_e^{-1} \otimes (\fweightreg{e}\mathbf{I})^{-1} ) \\
    & \propto \fweightreg{e}^{F_e D / 2} |\lprec_e|^{D/2}
              \exp( -\dfrac{1}{2} \fweightreg{e} \tr( \fweight{e} \lprec_e^{-1} \fweight{e}^\top ) )
\end{align}
where $\vecop(\fweight{e})$ is the vectorization of $\fweight{e}$, $\otimes$ denotes the Kronecker product and $\fweightreg{e} \geq 0$ is the diagonal element of the precision matrix.
The inclusion of $\lprec_e$ (the precision matrix of the latent vectors) in \eqref{eq:macau-betareg} is crucial for deriving a computationally efficient noise injection sampler, described in detail in Section~\ref{sec:macau-noise-injection}.

As the choice of $\fweightreg{e}$ is problem dependent, we set a gamma distribution as its hyperprior, as used in similar context for neural networks \cite{husmeier1999empirical}:
\begin{equation}
  \label{eq:macau-prec-hyperprior}
  p(\fweightreg{e} | \mu, \nu) = \mathcal{G}(\fweightreg{e} | \mu, \nu) 
                            \propto \fweightreg{e}^{\nu/2 - 1} \exp(- \dfrac{\nu}{2\mu} \fweightreg{e}),
\end{equation}
where $\mu$ and $\nu$ are fixed hyperparameters, which are both set to $1$ in the experiments.

\subsection{Relation Features}
Often there is extra information regarding observations (\emph{e.g.}, the day (from the release) when the user went to see the movie or the temperature of the chemical experiment).
However, this data is not linked to a single entity instance but instead to the observation (\emph{e.g.}, a particular user-movie pair).
If these features are fully observed for a particular relation $R$, then Macau can incorporate them directly into the observation model.
Let $\feat[R]{\jb} \in \mathbb{R}^{F_R}$ be the relation feature for observation $R_\jb$, then the previous observation model \eqref{eq:macau-noise} for relation $R$ is replaced by
\begin{equation}
  \label{eq:macau-relation-feature}
  p(R | \ub, \alpha_R) = \prod_{\jb \in I_R} \mathcal{N}(R_\jb | \mathbf{1} ^ \top \ub_\jb ^ {\eset_R} + \fweight{R}^\top \feat[R]{\jb}, \alpha_R^{-1}),
\end{equation}
where $\fweight{R} \in \mathbb{R}^{F_R}$ is the weight vector.
The treatment of $\fweight{R}$ is similar to $\fweight{e}$, \emph{i.e.}, Macau uses zero mean Gaussian prior on $\fweight{R}$ with precision $\fweightreg{R} \geq 0$ that has gamma hyperprior:
\begin{align}
  p(\fweight{R} | \fweightreg{R}) &= \mathcal{N}(\fweight{R} | \mathbf{0}, \fweightreg{R}\mathbf{I}) \\
  p(\fweightreg{R} | \mu, \nu)    &= \mathcal{G}(\fweightreg{R} | \mu, \nu),
\end{align}
where as before $\mu$ and $\nu$ are fixed hyperparameters, set to 1 in experiments.

\subsection{Gibbs Sampler}
\label{sec:gibbs}
Gibbs sampling is used to sample from the posterior of the model variables.
In this section we present the conditional distributions of the Gibbs sampler for all variables (except $\fweight{e}$ and $\fweight{R}$ for which we propose a specially designed sampler in Section~\ref{sec:macau-noise-injection}).

\subsubsection{Latent vectors}
Based on \eqref{eq:macau-latent-with-features} and \eqref{eq:macau-relation-feature} the conditional probability for $\latent{e}{i}$ is
\begin{align}
  p( \latent{e}{i} | \rset, \ub, \xb, \fweightall, \fweightregall, \alpha, \lprec_e )
  =& \mathcal{N}( \latent{e}{i} | \mu_i^{(e)*}, [\Lambda_i^{(e)*}]^{-1} )  
  \\
  \propto&
  \prod_{R \in \rset_e} \prod_{\jb \in \IR(e,i)} \mathcal{N}(R_\jb | \mathbf{1} ^ \top \ub_\jb ^ {\eset_R} + \fweight{R}^\top \feat[R]{\jb}, \alpha_R^{-1})
  \\
  & \times \mathcal{N}(\latent{e}{i} | \lmean_e + \fweight{e}^\top \feat[e]{i}, \lprec_e^{-1}),
  \nonumber
\end{align}
where
\begin{align*}
  \Lambda_i^{(e)*} &= \lprec_{e}
    + \sum_{R \in \rset_e} \alpha_{R} 
      \sum_{\jb \in \IR(e,i)} 
        \left(\dfrac{\mathbf{u}_\jb^{ \eset_R }}{\latent{e}{i}} \right)
        \left(\dfrac{\mathbf{u}_\jb^{ \eset_R }}{\latent{e}{i}} \right)^{\top}
  \\
  \mu_i^{(e)*}     &= [\Lambda_i^{(e)*}]^{-1}
  \left( 
    \lprec_e(\lmean_e + \fweight{e}^\top \feat[e]{i} )
    + \sum_{R \in \eset_R} \alpha_{R} 
      \sum_{\jb \in \IR(e,i)}
	 (R_\jb - \fweight{R}^\top \feat[R]{\jb})
         \dfrac{\mathbf{u}_\jb^{ \eset_R }}{ \latent{e}{i}}
    \right),
\end{align*}
where $\rset_e$ is the set of relations that the entity $e$ is linked to, $\mathbf{u}_\jb^{ \eset_R } / \latent{e}{i}$ denotes element-wise division\footnote{
The formula assumes that $\latent{e}{i}$ is only present once.
To handle such cases where, \emph{e.g.}, $\eset_R = (e,e)$, and there are observations on the diagonal, equations should be modified.
}, $\IR(e,i) \subseteq \IR$ is the set of indexes of observations to which instance $i$ of entity $e$ is linked to, \emph{i.e.}, all the observed data in relation $R$ for instance $i$.
The above equations use a shorthand that if relation $R$ does not have features then $\fweight{R}^\top \feat[R]{\jb}$ is $0$ and similarly if entity $e$ does not have features then $\fweight{e}^\top \feat[e]{i}$ is $0$.

\subsubsection{Gaussian priors}
Macau also uses the same Normal-Wishart hyperprior for $\lmean_e$ and $\lprec_e$ as BPMF \cite{salakhutdinov2008bayesian}:
\begin{equation}
  \label{eq:macau-hyperprior}
  p(\lmean_e, \lprec_e | \Theta_0) = 
    \mathcal{N}(\lmean_e | \lmean_0, (\beta_0 \lprec_e)^{-1})
    \mathcal{W}(\lprec_e | W_0, \nu_0),
\end{equation}
where the hyperparameters are set to uninformative values of $\lmean_0=\mathbf{0}$, $\beta_0 = 2$, $W_0 = \mathbf{I}$ (the identity matrix), and $\nu_0 = D$. Combining the hyperprior \eqref{eq:macau-hyperprior} with \eqref{eq:macau-latent-with-features} we get conditional probability
\begin{equation}
  \label{eq:macau-gibbs-prior}
  p(\lmean_e, \lprec_e | \ub^{(e)}, \xb, \fweightall, \Theta_0) = 
    \mathcal{N}(\lmean_e | \lmean^*_0, (\beta^*_0 \lprec_e)^{-1})
    \mathcal{W}(\lprec_e | W^*_0, \nu^*_0).
\end{equation}
Because of space constraints the formulas for $\lmean^*_0$, $\beta^*_0$, $W^*_0$, $\nu^*_0$ are presented in the Appendix~\ref{appendix:latent-prior}.
One of the essential differences compared to BPMF is that in BPMF the Gaussian priors model the latent vectors $\latent{e}{i}$ whereas in Macau they model the residual $\latent{e}{i} - \fweight{e}^\top \feat[e]{i}$.

\subsubsection{Precision parameter for the weight vector}
From \eqref{eq:macau-betareg} and \eqref{eq:macau-prec-hyperprior} we can derive the conditional probability for $\fweightreg{e}$ as
\begin{equation}
  p(\fweightreg{e} | \fweight{e}, \lprec_e, \mu, \nu) = \mathcal{G}(\fweightreg{e} | \tilde{\mu}, \tilde{\nu}),
\end{equation}
where
\begin{align*}
  \tilde{\nu} = F_e D + \nu
  \quad\quad\quad
  \tilde{\mu} = \dfrac{(F_e D + \nu) \mu}{\nu + \mu \tr(\fweight{e}^\top \fweight{e} \lprec_e)}.
\end{align*}
The conditional probability for $\fweightreg{R}$ is analogous and is described in Appendix~\ref{appendix:gibbs-for-weight-vector-of-relation-features}.

\subsection{Noise Injection Sampler}
\label{sec:macau-noise-injection}
From \eqref{eq:macau-latent-with-features} and \eqref{eq:macau-betareg} we can write out the conditional probability for $\fweight{e}$
\begin{align}
  &p(\fweight{e} | \lmean_e, \lprec_e, \ub, \xb, \fweightreg{e}) \\
  & \propto \exp\left( -\dfrac{1}{2} \sum_{i=1}^{N_e}
                                (\latent{e}{i} - \lmean_e - \fweight{e}^\top \feat[e]{i} )^\top
                                \lprec_e
                                (\latent{e}{i} - \lmean_e - \fweight{e}^\top \feat[e]{i} )
                       -\dfrac{1}{2} \fweightreg{e} \tr( \fweight{e} \lprec_e \fweight{e}^\top )
                \right)
  \nonumber
\end{align}
Let us denote
$\mathbf{U}=[\latent{e}{1}-\lmean_e,\ldots,\latent{e}{N_e}-\lmean_e]^{\top}$ and
$\Xb = [\feat[e]{1}, \ldots, \feat[e]{N}]^\top$ then, because both the likelihood and the prior contain $\lprec_e$, we can factorize $\lprec_e$ out:
\begin{equation}
  p(\fweight{e} | \lmean_e, \lprec_e, \ub, \xb, \fweightreg{e})
    \propto
    \exp\left(
      -\dfrac{1}{2}
       \tr[ ((\Ub - \Xb \fweight{e})^\top (\Ub - \Xb \fweight{e}) + \fweightreg{e}\fweight{e}^\top \fweight{e}) \lprec_e ]
    \right)
\end{equation}
and the Gaussian mean and precision can be derived (for details see Appendix~\ref{appendix:posterior-of-weight-vector}):
\begin{align}
  \label{eq:macau-gibbs-fweight}
  p(\fweight{e} | \lmean_e, \lprec_e, \ub, \xb, \fweightreg{e})
    \propto
    \exp \left(
      -\dfrac{1}{2}
       \vecop(\fweight{e} - \fweighthat{e})^\top
       (\lprec_e \otimes (\Xb^\top \Xb + \fweightreg{e} \mathbf{I}))
       \vecop(\fweight{e} - \fweighthat{e})
    \right)
\end{align}
where $\fweighthat{e} = (\Xb^\top \Xb + \fweightreg{e} \mathbf{I} )^{-1} \Xb^\top \Ub$ is the mean and $\lprec_e \otimes (\Xb^\top \Xb + \fweightreg{e} \mathbf{I})$ is the precision of the posterior.
However, even for moderate feature dimensions $F_e$ the standard sampling of the multivariate Gaussian is computationally \emph{intractable} because the size of the precision matrix is $DF_e \times DF_e$.

By exploiting the Kronecker product structure of the precision matrix and the existence of $(\Xb^\top \Xb + \fweightreg{e} \mathbf{I} )$ in both the mean and the precision we derive an alternative approach.
A sample of $\fweight{e}$ from \eqref{eq:macau-gibbs-fweight} can be obtained by solving a linear system for $\tilde{\beta}$:
\begin{equation}
  \label{eq:macau-gamblr}
  (\Xb^\top \Xb + \fweightreg{e}\mathbf{I}) \tilde{\beta} = \Xb^\top (\Ub + \Eb_1) + \sqrt{\fweightreg{e}} \Eb_2,
\end{equation}
where each row of matrices $\Eb_1 \in \mathbb{R}^{N_e \times D}$ and $\Eb_2 \in \mathbb{R}^{F_e \times D}$ is sampled from $\mathcal{N}(\mathbf{0}, \lprec_e^{-1})$.
The correctness of \eqref{eq:macau-gamblr} is proven in Appendix~\ref{appendix:correctness-of-noise-injection-sampler}.
The derivation of noise injection sampler for the weight vector $\fweight{R}$ for relation features is analogous, with the difference that the linear system has only single right-hand side.

Thus, to sample $\fweight{e}$ we need to solve a linear system of size $F_e \times F_e$ with $D$ different right-hand sides.
If $F_e$ is medium size (up to 20,000) we propose to use direct solvers\footnote{On a system with 2 Intel Xeon E5-2699 v3 CPUs using 8 cores Julia takes 25 seconds to solve a 20,000 $\times$ 20,000 system with 60 right-hand sides (= latent dimensions).}.
If $\Xb$ is sparse we can tackle high-dimensional systems by solving each right-hand side separately by using iterative method of conjugate gradient (CG).
CG only requires multiplication of $\Xb$ (and $\Xb^\top$) with a vector and can handle cases where $F_e$ is in the order of millions.

\section{Related Research}
\label{sec:related-research}
There are several extensions already proposed to BPMF that are related to our work.
Bayesian Probabilistic Tensor Factorization \cite{xiong2010temporal} extends BPMF to the factorization of a single 3-way relation (tensor) without entity and relation features.
Like BPMF their approach uses Gibbs sampler.

Singh and Gordon \cite{Singh2010relational} propose Bayesian MF method that can link together more than one 2-way relation (matrix).
Their sampling approach is analogous to BPMF except using Hamiltonian Monte Carlo within Gibbs where each latent vector is sampled separately by using Hamiltonian Monte Carlo.
Their method does not have support for tensors, entity features or relation features.
The lack of support for tensors also means their method cannot support multiple relations between two entities, which is useful, for example, in the case of drug-protein activity modeling (see Section~\ref{sec:chembl} for the experiment on ChEMBL data with two relations between potential drugs and proteins).

Hierarchical Bayesian Matrix Factorization with Side Information (HBMFSI) \cite{park2013hierarchical} is a method for the special case of factorizing a single matrix with entity features based on Variational Bayes.
HBMFSI does not allow the model to use relation features as in Macau.
However, they propose to add the concatenation of row entity features and column entity features in the same way as relation features in Macau, \emph{i.e.} $\feat[R]{i,j} = (\feat[row]{i}, \feat[col]{j})$.

\section{Experiments}
This section gives results for 1) the standard MF benchmark MovieLens and 2) a challenging biochemical problem based on the ChEMBL data set \cite{bento2014chembl}, and reports runtimes of Macau on a large-scale industrial drug--protein activity data set.
The performance reported is mean RMSE and the error bars in figures represent standard deviations.
All experiments are repeated 10 times.

\subsection{MovieLens Benchmark}

\begin{figure}[h]
\centering
\begin{minipage}[t]{.45\linewidth} 
\centering
\includegraphics[width=1\linewidth]{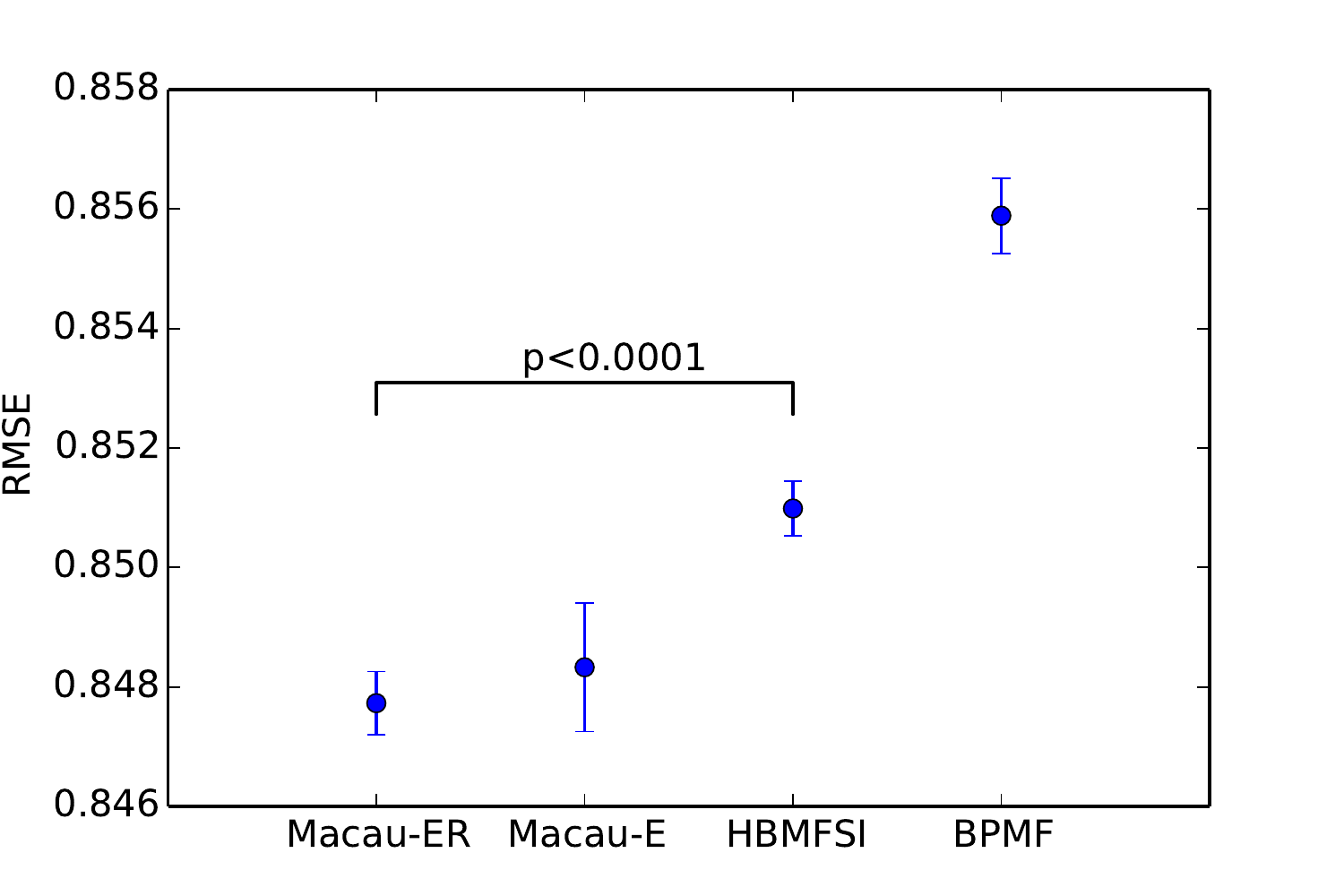}
\captionof{figure}{Results for MovieLens experiments. The p-value of the two-sided t-test between Macau-ER and HBMFSI is lower than 0.0001.}
\label{fig:movielens-rmse}
\end{minipage}
\hspace{1cm}
\begin{minipage}[t]{.45\linewidth}
\centering
\includegraphics[width=1\linewidth]{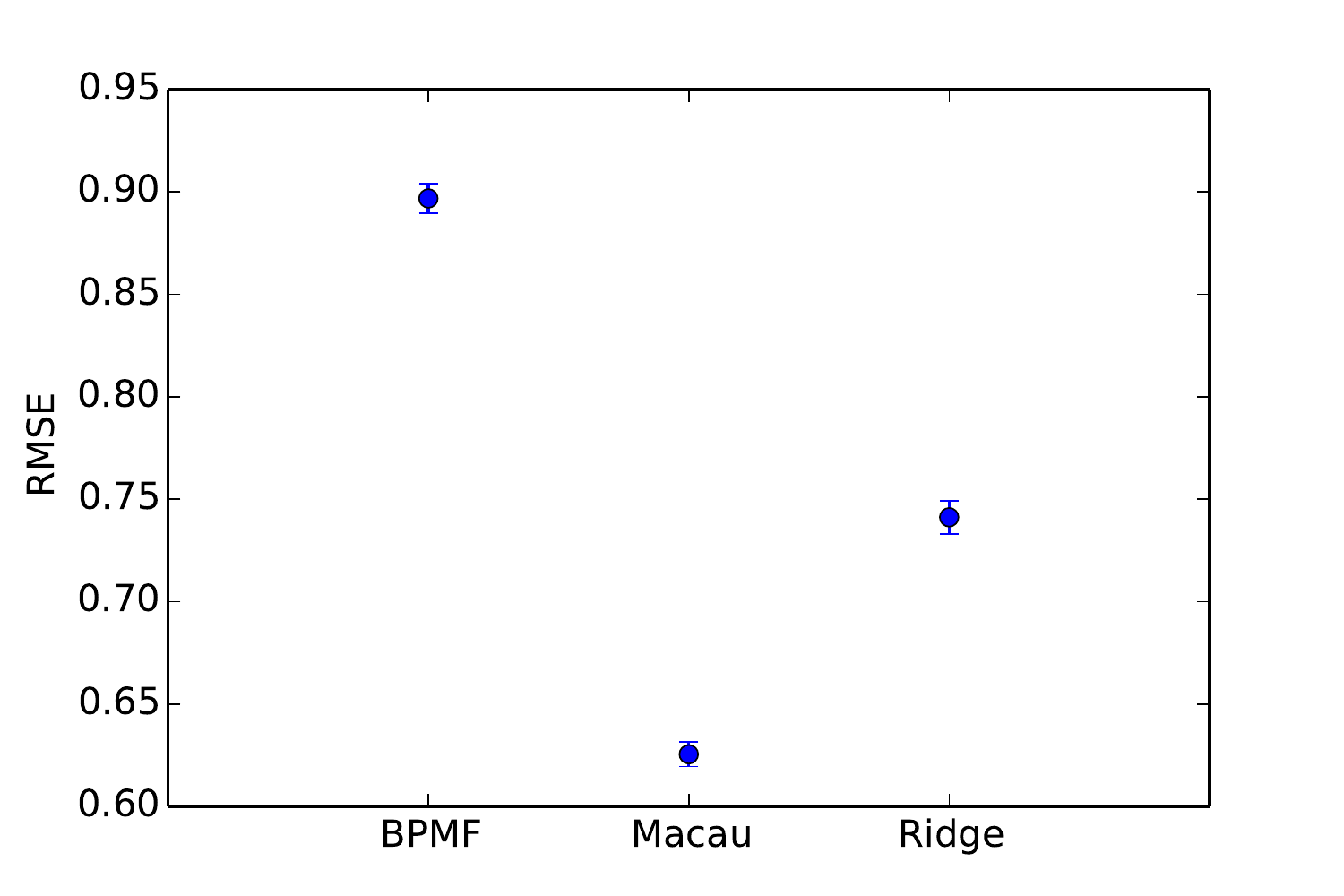}
\captionof{figure}{Results for ChEMBL experiments. BPMF and Macau use $D=30$.}
\label{fig:chembl-rmse}
\end{minipage}
\end{figure}

The MovieLens data set consists of a single matrix of movie--user ratings from 6,040 users and 3,952 movies.
There are total of 1,000,209 ratings taking values from 1.0 to 5.0.
Recent research \cite{amatriain2009like} has investigated in detail the noise level in movie ratings.
Amatriain \emph{et al.} made a conservative estimate of between-trial RMSE of $0.8156$.
Based on that estimate, we chose to use $\alpha=1.5$ in our experiments.
The data set contains 29 and 18 dimensional entity features for users and movies, respectively.
We compare Macau against HBMFSI\footnote{In the experiments we used the MATLAB implementation of HBMFSI provided by the authors.}, which is the state-of-the-art MF approach with entity features, using the same evaluation setup used in their paper \cite{park2013hierarchical}.
Namely, one half of the ratings are randomly set as the test set and another half as the training set.
The methods compared are 1) Macau-E: Macau with entity features, 2) Macau-ER: Macau with entity features and relation features constructed as in HBMFSI (see Section~\ref{sec:related-research}), 3) HBMFSI, and 4) BPMF.
All methods use latent dimension $D=30$.
It should be noted that the relative performance between the methods was similar when we used $D=10$ (data not shown).
The results show that both Macau setups outperform HBMFSI and BPMF, see Figure~\ref{fig:movielens-rmse}.

\subsection{ChEMBL drug--protein activity prediction}
\label{sec:chembl}
The prediction of drug and protein interactions is crucial for the development of new drugs.
In this case study we focus on the interaction measure IC50, which measures the concentration of the drug necessary to inhibit the activity of the protein by $50\%$.
We prepared a data set from the public bioactivity database ChEMBL\cite{bento2014chembl} Version 19.
First, we selected proteins that had at least 200 IC50 measurements, and then we kept drugs with 3 or more IC50 measurements.
Finally, we filtered out some measurements with clear data errors (these were also reported to ChEMBL).
The final numbers for small molecules and proteins are 15,073 and 346, respectively, with total of 59,280 IC50 measurements.
In all of the ChEMBL experiments we model $\log_{10}$ of IC50 and set $\alpha=5.0$, because this corresponds to a reasonable standard deviation of $0.45$\footnote{It is possible to enhance the performance by tuning/sampling $\alpha$.}.
For drugs, we use sparse features (substructure fingerprint) with $F_{drug}=\text{105,672}$, for proteins, we use dense features (based on protein sequence) with $F_{prot}=\text{20}$.

In the first experiment, we compare Macau with entity features for drugs and proteins to BPMF, as well as individual  ridge regression based on drug features, for each protein.
Macau and BPMF use 30 latent dimensions, because we observed it is sufficient for good performance, see Appendix~\ref{appendix:chembl-latent-dimension}.
To tune the regularization parameter of ridge regression of each protein, we used 5-fold inner cross-validation.
A test set containing 20\% of the observations is chosen at random.  
The strong performance of Macau over BPMF, as seen in Figure~\ref{fig:chembl-rmse}, is expected because Macau gives the most advantage when the relation is sparsely observed.

\begin{figure}[h]
\centering
\begin{minipage}[t]{.45\linewidth}
\centering
\includegraphics[width=0.9\linewidth]{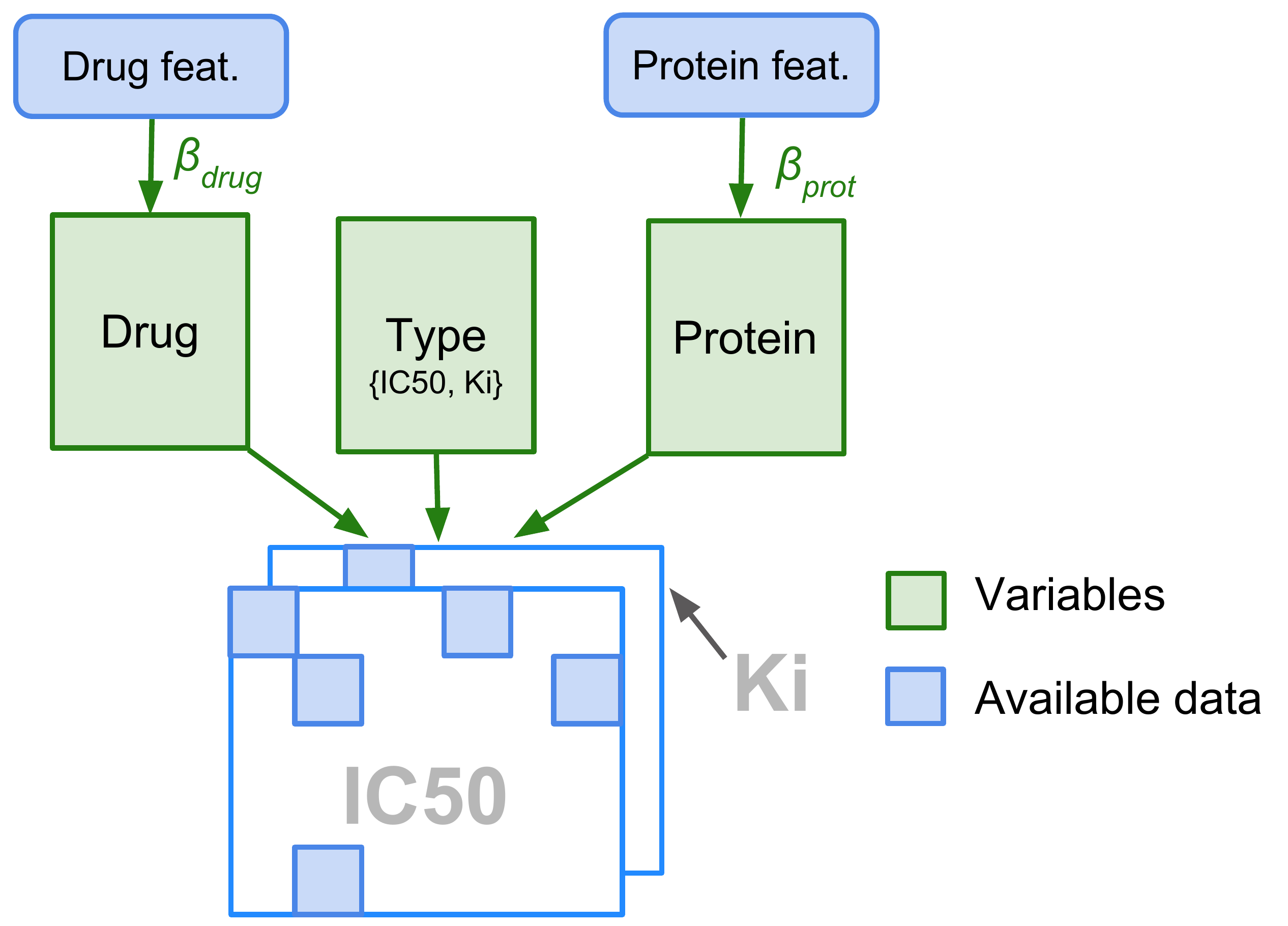}
\captionof{figure}{The IC50+Ki model has two \emph{Type}s of relations between entity \emph{Drug} and entity \emph{Protein}.}
\label{fig:tensor-model}
\end{minipage}
\hspace{1cm}
\begin{minipage}[t]{.45\linewidth}
\centering
\includegraphics[width=1\linewidth]{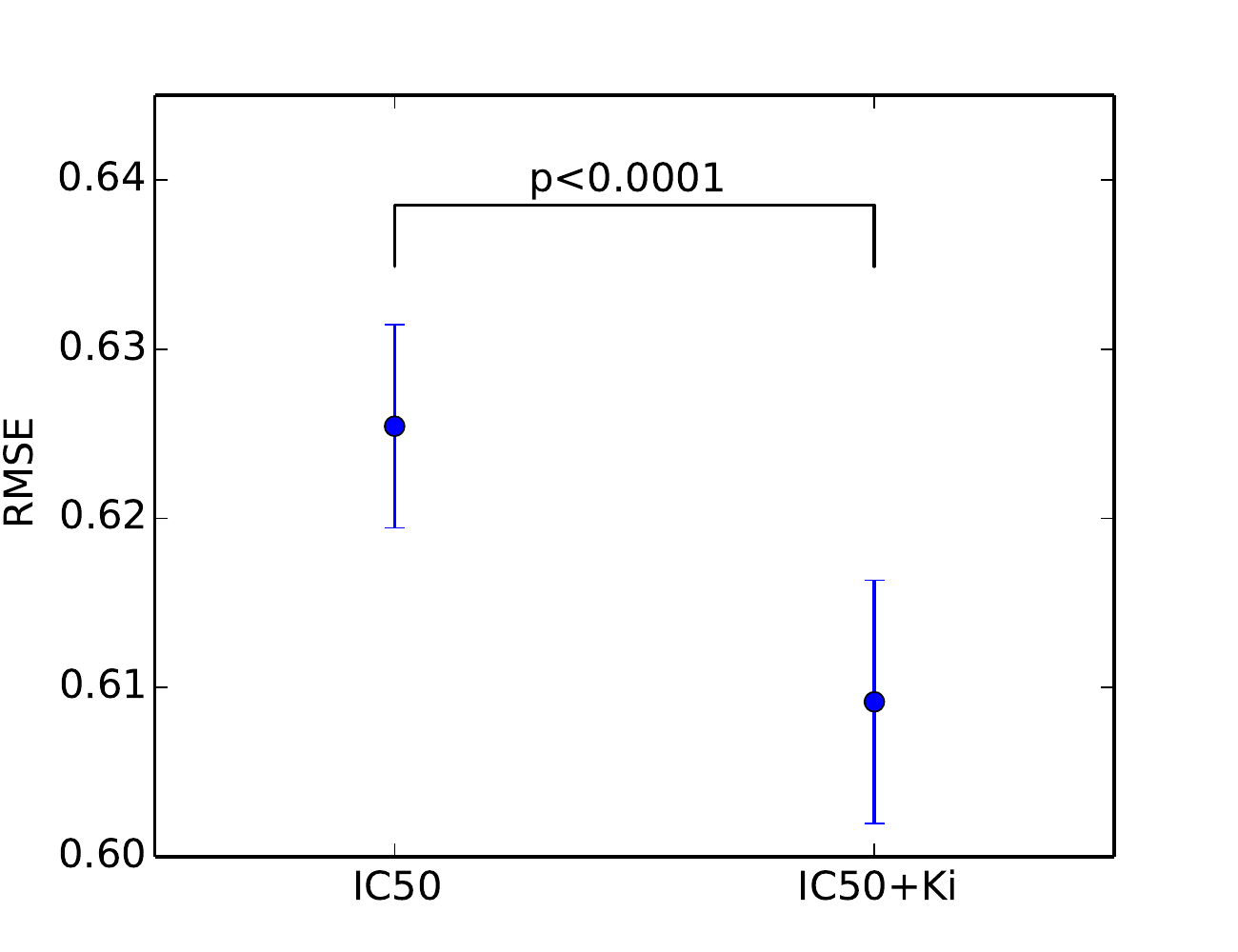}
\captionof{figure}{Comparison of IC50 and IC50+Ki Macau models using $D=30$.}
\label{fig:tensor-rmse}
\end{minipage}
\end{figure}

In the second experiment we want to improve IC50 predictions by introducing a new relation, Ki, between drugs and proteins.
Ki measures the binding affinity of the drug for a protein. While related to IC50, it measures a different biochemical aspect of the interaction. 
We thus expect that it contributes additional information for our task.
In Macau, multiple relations between two entities are represented as a tensor by creating a third entity denoting the type of interaction,
see Figure~\ref{fig:tensor-model}. The dimensions of the tensor are $ \text{15,073} \times 346 \times 2$, and the Ki part contains 5,121 observations.
As before the test set contains measurements only from the IC50 part. 
As can be seen from Figure~\ref{fig:tensor-rmse} the tensor model IC50+Ki significantly outperforms the single relation model with only IC50 ($p<0.0001$).

\begin{figure}[t] 
\centering
\begin{minipage}[t]{.45\linewidth}
\centering
\includegraphics[width=0.9\linewidth]{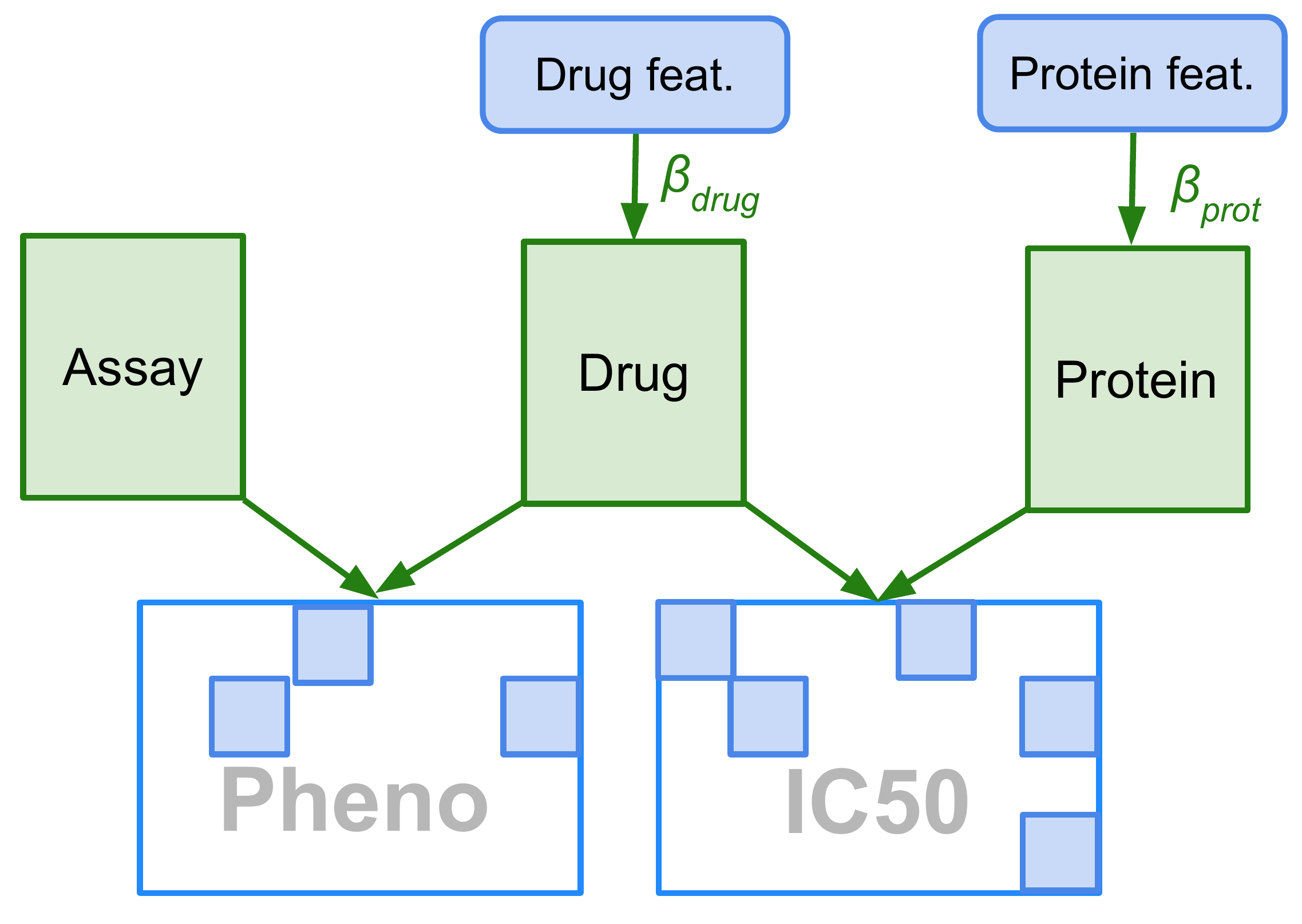}
\captionof{figure}{The IC50+Pheno model has three entities and two relations.}
\label{fig:pheno-model}
\end{minipage}
\hspace{1cm}
\begin{minipage}[t]{.45\linewidth}
\centering
\includegraphics[width=1\linewidth]{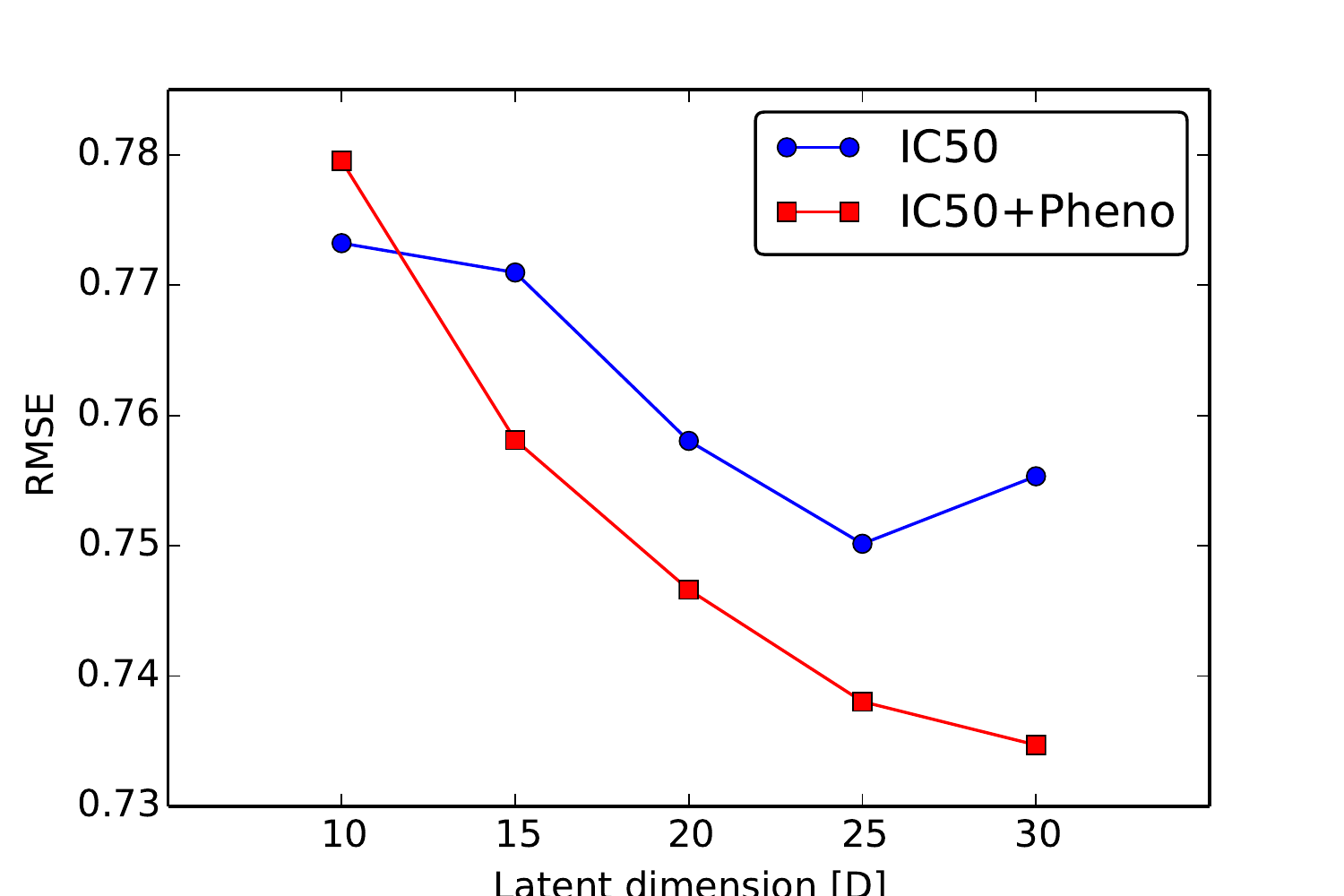}
\captionof{figure}{Comparison of IC50 and IC50+Pheno models.}
\label{fig:pheno-rmse}
\end{minipage}
\end{figure}

In the final experiment, we explore the effect of connecting drugs to two relations. 
For the drugs in our IC50 data set, we compiled all cancer assays from ChEMBL that had at least 20 compounds in that set.
From these, we created a new relation \emph{Pheno} with $7,552$ observations measuring the phenotypic effect of drugs in $182$ \emph{Assay}s, which is depicted in Figure~\ref{fig:pheno-model}.
Because the effects of assays are not directly linked to any specific protein, we expected weaker effect than from the Ki data.
Therefore, for the comparison of IC50 and IC50+Pheno models the $20\%$ test set of IC50 measurements are selected only from that 1,678 compounds that have \emph{Pheno} observations.
In Figure~\ref{fig:pheno-rmse} we can observe the IC50+Pheno model outperforms the IC50 model when appropriately large latent dimension is used.
It is interesting to note, that with $D=10$ the IC50+Pheno model is slightly losing in performance, which can be an evidence that, with too small $D$, adding more relations can result in an overcrowded latent space.
\subsection{Runtime on Large-scale Industrial Data Set}
For large scale problems, our implementation has multi-core and multi-node parallelization.
The sampling of latent vectors can be parallelized straightforwardly as the the latent vectors of a single entity are, in our use cases, independent of each other and can be sampled in parallel.
The only difference here compared to BPMF is that, for entities that have features Macau requires computing $\fweight{e}^\top \feat[e]{i}$, for which our implementation provides parallelization as well.

In the parallelization of \eqref{eq:macau-gamblr}, if $F_e \leq 20000$, the direct solver is fast enough not to require additional parallelization.
As mentioned in the case of sparse features, Macau uses CG to solve \eqref{eq:macau-gamblr} for each right-hand side separately.
For each CG, our implementation parallelizes the matrix product operations in a multi-core way and CGs can be distributed across multiple processes and thus can be parallelized over multiple nodes.

The large-scale data set is a subset of a proprietary data set from Janssen Pharmaceutica containing millions of compounds.
The subset has more than 1.8M compounds and more than 1,000 proteins for a total of several tens of millions of compound--protein measurements.
Here we report the computation times of Macau on two types of features for the compounds using systems with 2 Intel Xeon E5-2699 v3 CPUs.
Firstly, for the feature dimension $F_e \approx 6000$ and $D=30$, the computation of the full Gibbs step using 8 cores of a single node takes about 40 seconds (using a direct solver for \eqref{eq:macau-gamblr}).
Secondly, for the feature dimension $F_e \approx \text{4,000,000}$ having sparsity $0.002\%$ and $D=30$ the computation of the full Gibbs step using 10 cores per CG and total of 15 nodes (2 CGs per node) takes about 600 seconds.
We observed that 1,000 Gibbs iterations (from which 800 were discarded as burn-in) were sufficient to reach a stable posterior.

\section{Conclusion}
The best of our knowledge, this paper proposes the first Bayesian factorization method that allows to handle tensors, multiple relations, and entity and relation features.

\subsubsection*{Acknowledgments}
Jaak Simm, Adam Arany, Pooya Zakeri and Yves Moreau are funded by Research Council KU Leuven (CoE PFV/10/016 SymBioSys) and by Flemish Government (IOF, Hercules Stitching, iMinds Medical Information Technologies SBO 2015, IWT: O\&O ExaScience Life Pharma; ChemBioBridge, PhD grants).

\subsubsection*{References}
\bibliographystyle{abbrv}

\begingroup
\renewcommand{\section}[2]{}%
\bibliography{ref}
\endgroup

\appendix
\section{Appendices}

\subsection{Data Models Supported by Macau}
\label{appendix:data-models}

Let $M$ be a Macau model (\emph{i.e.}, hypergraph) specifying the entities and their relations.
We say $M$ is \emph{factorizable} if given large enough latent dimension $D$ we can choose latent vectors for every entity in such a way that they can fit any relation data with arbitrarily small error.

However, it is clear that some models are not \emph{factorizable}. For example, consider two entities $e_1$ and $e_2$, with the latent vectors $\ub$ and $\vb$, respectively, and two relations, $R$ and $S$, between them.
Since both relations are modelled by the same formula $\ub_i ^\top \vb_j$ it is not possible to fit arbitrary data.
Actually we can only fit the case when the two relations are equal (\emph{i.e.,} $R_{ij} = S_{ij}$).

Let us define the relation $R$ in model $M$ to be \emph{factorizable} if in the single latent dimensional case ($D = 1$) it is possible to specify arbitrary values to the latent variables of its entities, $e \in \eset_R$, while keeping the predictions for all other relations in the model $M$ equal to $0$.
It is straight-forward to see that if $R$ is \emph{factorizable} we can fit any observed data of the relation $R$ by adding new latent dimensions without affecting the predictions for other relations.
Additionally, the fact that all relations of $M$ are \emph{factorizable} implies $M$ is \emph{factorizable}, because we can always add new latent dimensions that only effect a specific relation and, thus, fit all relations as accurately as needed.

It is easy to show that if all pairs of entities in the hypergraph (Macau model) $M$ have at most one hyperedge (relation) between them, the model $M$ is \emph{factorizable}.
To see that this is true consider a relation $R$ in such a model.
$R$ is factorizable because if we set the latent variables of the non-participating entities, $e \notin \eset_R$, equal to $0$ then the predictions of the other relations will be zero as they all contain at least one non-participating entity. 

From this we can see that Macau can factorize any
\begin{itemize}
 \item ordinary undirected graph,
 \item acyclic hypergraph.
\end{itemize}
Additionally, it is also possible to tensorize simple cases when there are multiple edges between two entities. 
For example, the model IC50+Ki in our paper has two relations, namely IC50 and Ki, between the entities Drug and Protein.
To handle it in Macau we represent it as a tensor with three modes: Drug, Protein, Type, where the third mode specifies the relation (either IC50 or Ki).

\subsection{Sampling of Gaussian Priors of the Latent Variables}
\label{appendix:latent-prior}
In Macau $\lmean_e$ and $\lprec_e$ are interpreted as the model for the residuals $\latent{e}{i} - \fweight{e}^\top \feat[e]{i}$.
The conditional joint probability $\lmean_e$, $\lprec_e$ used in sampler is
\begin{equation}
  p(\lmean_e, \lprec_e | \ub^{(e)}, \xb, \fweightall, \Theta_0) = 
    \mathcal{N}(\lmean_e | \lmean^*_0, (\beta^*_0 \lprec_e)^{-1})
    \mathcal{W}(\lprec_e | W^*_0, \nu^*_0),
\end{equation}
where
\begin{align}
 \lmean^*_0 &= \dfrac{\beta_0 \lmean_0 + N_e \bar{U}}{\beta_0 + N_e},\\
 \beta^*_0  &= \beta_0 + N_e,\\
 \label{eq:app:nustar}
 \nu^*_0    &= \nu_0 + N_e + F_e,\\
 \label{eq:app:Wstar}
 [W^*_0]^{-1} &= W_0^{-1} + N_e\bar{S}
    + \beta_0 \lmean_0 \lmean_0^\top
    - \beta_0^* \lmean_0^* {\lmean_0^*}^\top
    + \fweightreg{e} \fweight{e}^\top \fweight{e},
  \\
 \bar{U} &= \dfrac{1}{N_e} \sum_{i=1}^{N_e} (\latent{e}{i} - \fweight{e}^\top \feat[e]{i}),\\
 \bar{S} &= \dfrac{1}{N_e} \sum_{i=1}^{N_e} (\latent{e}{i} - \fweight{e}^\top \feat[e]{i}) (\latent{e}{i} - \fweight{e}^\top \feat[e]{i})^\top,
\end{align}
where $N_e$ is the number of instances of entity $e$.
Note that the terms $\fweightreg{e} \fweight{e}^\top \fweight{e}$ in \eqref{eq:app:Wstar} and $F_e$ in \eqref{eq:app:nustar} come due to the dependence of the prior of $\fweight{e}$ on $\lprec_e$, see \eqref{eq:macau-betareg}.

\subsection{Gibbs Sampling of Precision Parameter of Weight Vector of Relation Features}
\label{appendix:gibbs-for-weight-vector-of-relation-features}
Recall that the $\fweightreg{R}$ is the diagonal value for the precision variable in the prior for weight vector $\fweight{R}$ and that the hyperprior of $\fweightreg{R}$ is gamma distribution with fixed parameters $\mu$ and $\nu$:
\begin{align}
  p(\fweight{R} | \fweightreg{R}) &= \mathcal    {N}(\fweight{R} | \mathbf{0}, \fweightreg{R}\mathbf{I}) \\
  p(\fweightreg{R} | \mu, \nu)    &= \mathcal{G}(\fweightreg{R} | \mu, \nu),
\end{align}
where
\begin{equation}
  \mathcal{G}(\fweightreg{R} | \mu, \nu) \propto 
    \fweightreg{R}^{\nu/2 - 1} \exp(- \dfrac{\nu}{2\mu} \fweightreg{R}).
\end{equation}
The conditional probability is then
\begin{align}
  p(\fweightreg{R} | \fweight{R}, \mu, \nu)
  & \propto p(\fweight{R} | \fweightreg{R}) p(\fweightreg{R} | \mu, \nu) \\
  & \propto \fweightreg{R}^{F_R / 2}
            \exp\left(- \dfrac{1}{2} \fweightreg{R} \fweight{R}^\top \fweight{R}  \right)
            \fweightreg{R}^{\nu/2 - 1}
            \exp\left(- \dfrac{\nu}{2\mu} \fweightreg{R} \right) \\
  & \propto \fweightreg{R}^{(F_R + \nu)/2 - 1} 
            \exp\left(- \fweightreg{R}\dfrac{ \nu + \mu \fweight{R}^\top \fweight{R}}{2\mu}    \right) \\
  & \propto \fweightreg{R}^{(F_R + \nu)/2 - 1} 
            \exp\left(- \fweightreg{R}\dfrac{ (\nu + \mu \fweight{R}^\top \fweight{R})(F_R + \nu)}{2(F_R + \nu)\mu}    \right) \\
  & \propto \mathcal{G}(\fweightreg{R} | \tilde{\mu}, \tilde{\nu}),
\end{align}
where
\begin{align}
  \tilde{\nu} & = F_R + \nu,
  \\
  \tilde{\mu} & = \dfrac{(F_R + \nu) \mu}{\nu + \mu \fweight{R}^\top \fweight{R}},
\end{align}
where $F_R$ is the dimensionality of the relation features of $R$.

\subsection{Derivation of Gaussian Mean and Precision for the Weight Vector of the Entity Features}
\label{appendix:posterior-of-weight-vector}
The conditional probability for $\fweight{e}$ is
\begin{align}
  & p(\fweight{e} | \lmean_e, \lprec_e, \ub, \xb, \fweightreg{e})
  \\
  \label{eq:sup-conditional-fweight}
  & \propto
    \exp\left(
      -\dfrac{1}{2}
       \tr[ ((\Ub - \Xb \fweight{e})^\top (\Ub - \Xb \fweight{e}) + \fweightreg{e}\fweight{e}^\top \fweight{e}) \lprec_e ]
    \right).
\end{align}
Next we use the link $\Xb^\top \Ub = (\Xb^\top \Xb + \fweightreg{e} \mathbf{I}) \fweighthat{e}$ (from the definition of $\fweighthat{e}$) and expand the inner part of \eqref{eq:sup-conditional-fweight}:
\begin{align}
  & (\Ub - \Xb \fweight{e})^\top (\Ub - \Xb \fweight{e}) + \fweightreg{e}\fweight{e}^\top \fweight{e} \\
  & = \Ub^\top \Ub
    + \fweight{e}^\top \Xb^\top \Xb \fweight{e} 
    - \Ub^\top \Xb \fweight{e} - \fweight{e}^\top \Xb^\top \Ub
    + \fweightreg{e}\fweight{e}^\top \fweight{e}.
  \\
  & = \Ub^\top \Ub
    + \fweight{e}^\top (\Xb^\top \Xb + \fweightreg{e}\mathbf{I})\fweight{e}
    - \Ub^\top \Xb \fweight{e} - \fweight{e}^\top \Xb^\top \Ub
  \\
  & = \Ub^\top \Ub
    + \fweight{e}^\top    (\Xb^\top \Xb + \fweightreg{e} \mathbf{I}) \fweight{e}
    - \fweighthat{e}^\top (\Xb^\top \Xb + \fweightreg{e} \mathbf{I}) \fweight{e}
    - \fweight{e}^\top    (\Xb^\top \Xb + \fweightreg{e} \mathbf{I}) \fweighthat{e}
  \\
  & = \Ub^\top \Ub
    + \fweight{e}^\top    (\Xb^\top \Xb + \fweightreg{e} \mathbf{I}) \fweight{e}
    - \fweighthat{e}^\top (\Xb^\top \Xb + \fweightreg{e} \mathbf{I}) \fweight{e}
    - \fweight{e}^\top    (\Xb^\top \Xb + \fweightreg{e} \mathbf{I}) \fweighthat{e}
    \nonumber
  \\ 
    & \quad
    + \fweighthat{e}^\top (\Xb^\top \Xb + \fweightreg{e} \mathbf{I}) \fweighthat{e}
    - \fweighthat{e}^\top (\Xb^\top \Xb + \fweightreg{e} \mathbf{I}) \fweighthat{e}
  \\
  & = \underbrace{\Ub^\top \Ub}_{\text{const}}
    + (\fweight{e}^\top - \fweighthat{e})^\top
      (\Xb^\top \Xb + \fweightreg{e} \mathbf{I})
      (\fweight{e}^\top - \fweighthat{e})
    - \underbrace{\fweighthat{e}^\top (\Xb^\top \Xb + \fweightreg{e} \mathbf{I}) \fweighthat{e}}_{\text{const}}
\end{align}
Next we plug the non-constant part back to \eqref{eq:sup-conditional-fweight}
\begin{align}
  & p(\fweight{e} | \lmean_e, \lprec_e, \ub, \xb, \fweightreg{e})
  \\
  & \propto
    \exp\left(
      -\dfrac{1}{2}
       \tr[ (
	(\fweight{e}^\top - \fweighthat{e})^\top
	(\Xb^\top \Xb + \fweightreg{e} \mathbf{I})
	(\fweight{e}^\top - \fweighthat{e})
      ) \lprec_e ]
    \right).
  \\
  & \propto
    \exp \left(
      -\dfrac{1}{2}
       \vecop(\fweight{e} - \fweighthat{e})^\top
       (\lprec_e \otimes (\Xb^\top \Xb + \fweightreg{e} \mathbf{I}))
       \vecop(\fweight{e} - \fweighthat{e})
    \right),
\end{align}
where we can clearly see the precision and mean of $\fweight{e}$.

\subsection{Correctness of Noise Injection Sampler}
\label{appendix:correctness-of-noise-injection-sampler}
Let $\Xb$, $\Ub$, $\lprec_e$ be matrices described in Gibbs sampling section of $\fweight{e}$.
Here we prove a more general version of the sampler where instead of precision matrix $\fweightreg{e} \mathbf{I}$ we allow any positive definite matrix $\boldsymbol{\Lambda} \in \mathbb{R}^{F \times F}$.
Let $\sqrt{\Lambdab}$ be a matrix such that $\sqrt{\Lambdab} \sqrt{\Lambdab}^\top = \Lambdab$.

\begin{lemma}
Let $\Eb_1 \in \mathbb{R}^{N_e \times D}$ and $\Eb_2 \in \mathbb{R}^{F_e \times D}$ be matrices where their each row is independently generated from $\mathcal{N}(\mathbf{0}, \lprec_e^{-1})$ and let the variable $\tilde{\beta}$ be the solution to the linear system
\begin{equation}
 (\Xb^\top \Xb + \Lambdab) \tilde{\beta} = \Xb^\top (\Ub + \Eb_1) + \sqrt{\Lambdab} \Eb_2,
\end{equation}
then $\vecop(\tilde{\beta})$ is distributed by multinomial Gaussian distribution with mean $\vecop((\Xb^\top \Xb + \Lambdab )^{-1} \Xb^\top \Ub)$ and precision $\lprec_e \otimes (\Xb^\top \Xb + \Lambdab)$.
\end{lemma}

\begin{proof}
From
\begin{equation}
  \tilde{\beta} = (\Xb^\top \Xb + \Lambdab)^{-1} (\Xb^\top (\Ub + \Eb_1) + \sqrt{\Lambdab} \Eb_2)
\end{equation}
it is clear that $\tilde{\beta}$ is distributed by Gaussian as it constructed by affine transformations and sums of Gaussian variables. As $\Eb_1$ and $\Eb_2$ have zero mean we get
\begin{align}
  \mathbb{E}[\tilde{\beta}]
  & = (\Xb^\top \Xb + \Lambdab)^{-1}( \Xb^\top \mathbb{E}[\Ub + \Eb_1] + \sqrt{\Lambdab} \mathbb{E}[ \Eb_2 ] )
  \\
  & = (\Xb^\top \Xb + \Lambdab)^{-1} \Xb^\top \Ub,
\end{align}
proving the correctness of the mean.

For the precision we investigate the covariance between $i$ and $j$ column of $\tilde{\beta}$. In what follows we use notation $\mathbf{A}_i$ to denote the column $i$ of matrix $\mathbf{A}$. Let's also denote $\Kb = (\Xb^\top \Xb + \Lambdab)^{-1}$, giving us $\mathbb{E}[\tilde{\beta}] = \Kb \Xb^\top \Ub$, then
\begin{align}
\operatorname{cov}(\tilde{\beta}_i, \tilde{\beta}_j)
& = \mathbb{E}\left[ (\tilde{\beta}_i - \mathbb{E}[\tilde{\beta}_i]) (\tilde{\beta}_j - \mathbb{E}[\tilde{\beta}_j])^\top \right] \\
& = \mathbb{E}
    \left[
      \left(\Kb(\Xb^\top (\Ub + \Eb_1)_i + (\sqrt{\Lambdab} \Eb_2)_i) - \Kb \Xb^\top \Ub_i   \right)
      \right.
      \\
&\quad\quad\,\,
      \left.
      \cdot
      \left(\Kb(\Xb^\top (\Ub + \Eb_1)_j + (\sqrt{\Lambdab} \Eb_2)_j) - \Kb \Xb^\top \Ub_j   \right)^\top
    \right]
    \nonumber
\\
& = \Kb \mathbb{E}
    \left[
      \left( (\Xb^\top \Eb_1)_i + (\sqrt{\Lambdab} \Eb_2)_i \right)
      \left( (\Xb^\top \Eb_1)_j + (\sqrt{\Lambdab} \Eb_2)_j \right) ^ \top
    \right]
    \Kb
\\
\label{eq:noise-cov-expansion}
& =       \Kb \Xb^\top \mathbb{E} \left[ (\Eb_1)_i ((\Eb_1)_j)^\top \right] \Xb \Kb \\
& \quad + \Kb \Xb^\top \mathbb{E} \left[ (\Eb_1)_i ((\Eb_2)_j)^\top \right] \sqrt{\Lambdab}^\top \Kb \nonumber \\
& \quad + \Kb \sqrt{\Lambdab} \mathbb{E} \left[ (\Eb_2)_i ((\Eb_1)_j)^\top \right] \Xb \Kb \nonumber \\
& \quad + \Kb \sqrt{\Lambdab} \mathbb{E} \left[ (\Eb_2)_i ((\Eb_2)_j)^\top \right] \sqrt{\Lambdab}^\top \Kb. \nonumber
\end{align}
The expectations in the first and last term of the equation~\eqref{eq:noise-cov-expansion} give
\begin{align}
  \mathbb{E} [ (\Eb_1)_i ((\Eb_1)_j)^\top ] & = (\lprec_e^{-1})_{i,j} \mathbf{I}_{N_e}
  \\
  \mathbb{E} [ (\Eb_2)_i ((\Eb_2)_j)^\top ] & = (\lprec_e^{-1})_{i,j} \mathbf{I}_{F_e},
\end{align}
where $\mathbf{I}_n$ is $n$-dimensional identity matrix.
The middle two terms are equal to zero because $\mathbb{E}\left[(\Eb_{2})_{i}(\Eb_{1})_{j}^{\top}\right]=\mathbf{0}$ due to $\Eb_1$ and $\Eb_2$ being zero mean and independent of each other.
Thus, we get
\begin{align}
\operatorname{cov}(\tilde{\beta}_i, \tilde{\beta}_j)
& = \Kb \Xb^\top (\lprec_e^{-1})_{i,j} \Xb \Kb + \Kb \sqrt{\Lambdab} (\lprec_e^{-1})_{i,j} \sqrt{\Lambdab}^\top \Kb \\
& = (\lprec_e^{-1})_{i,j} \Kb(\Xb ^ \top \Xb + \Lambdab) \Kb
\\
& = (\lprec_e^{-1})_{i,j} \Kb
\\
& = (\lprec_e^{-1})_{i,j} (\Xb^\top \Xb + \Lambdab)^{-1}.
\end{align}
This means the covariance matrix of $\vecop(\tilde{\beta})$ is $\lprec_e^{-1} \otimes (\Xb^\top \Xb + \Lambdab)^{-1}$ and thus the precision is $\lprec_e \otimes (\Xb^\top \Xb + \Lambdab)$.
\end{proof}

\subsection{Macau Performance with Different Latent Dimensions in ChEMBL}
\label{appendix:chembl-latent-dimension}
\begin{figure}[h]
\centering
\includegraphics[width=0.6\linewidth]{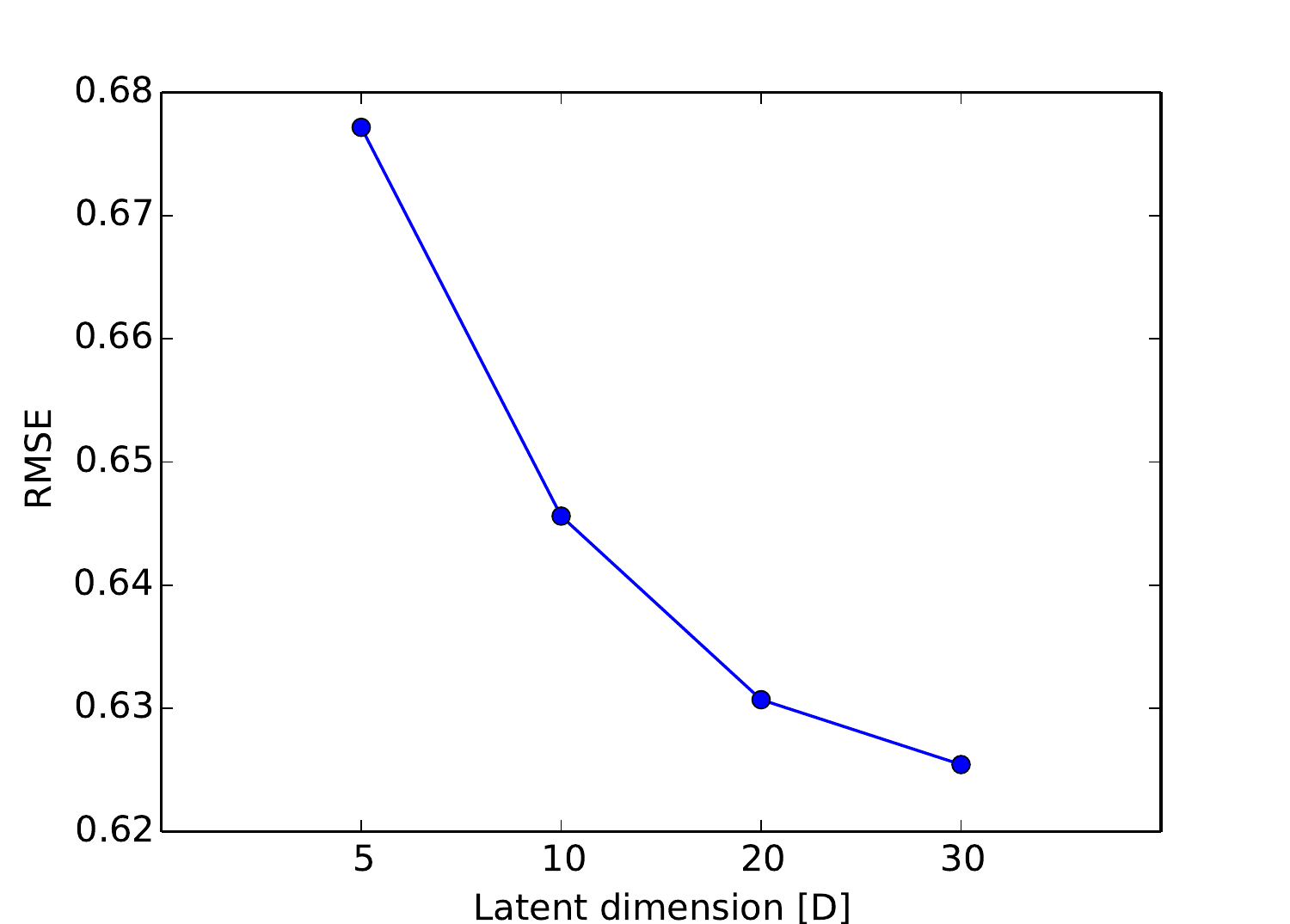}
\caption{The IC50 model with different latent dimensions.}
\label{fig:macau-dimension}
\end{figure}
Figure~\ref{fig:macau-dimension} shows the effect of the number of latent dimensions on performance of Macau on the model where a matrix of IC50 observations are factorized using entity features on drugs and proteins.

\end{document}